\pdfoutput=1
\documentclass[11pt]{article}

\usepackage[usenames,dvipsnames]{xcolor}
\usepackage{color}
\usepackage{times}
\usepackage{epsfig}
\usepackage{graphicx}
\usepackage{amsmath}
\usepackage{amsthm}
\usepackage{amssymb}
\usepackage{mathtools}
\usepackage{multirow}
\usepackage{bbm}
\usepackage{dsfont}
\newcommand{\cut}[1]{}

\newcommand{\loss}{\mathcal{L}}

\newcommand{\postspace}{\vskip -3mm}
\newcommand{\minipostspace}{\vskip -2.5mm}

\newcommand{\gradloss}{\nabla_{\theta}{\mathcal{L}}}

\usepackage{naacl2021}%\usepackage[review]{naacl2021}

\usepackage{times}
\usepackage{latexsym}
\usepackage{booktabs,siunitx}
\usepackage{enumitem}
\usepackage{soul}
\usepackage[T1]{fontenc}
\usepackage[utf8]{inputenc}

\usepackage{microtype}

\usepackage{subcaption}

\title{An Empirical Comparison of Instance Attribution Methods for NLP}

\author{Pouya Pezeshkpour\thanks{Equal contribution}\\
  University of California, Irvine \\
  \texttt{pezeshkp@uci.edu} \\ \And
Sarthak Jain\textcolor{darkblue}{$^*$}\\%$^*$
  Northeastern University \\
  \texttt{jain.sar@northeastern.edu} \\
\AND
Byron C. Wallace \\
  Northeastern University \\
  \texttt{b.wallace@northeastern.edu}\\
  \And
 Sameer Singh \\
  University of California, Irvine \\
  \texttt{sameer@uci.edu}
\\}

\begin{document}
\maketitle
\begin{abstract}
%With the 
Widespread adoption of deep models has motivated a pressing need for approaches to interpret network outputs and to facilitate model debugging.
%play a major role in assessing trust. 
%A family of attribution methods provide explanation through computing influence of training samples for target instance prediction. 
\emph{Instance attribution} methods constitute one means of accomplishing these goals by retrieving training instances that (may have) led to a particular prediction.
Influence functions (IF; \citealt{koh2017understanding}) provide machinery for doing this by quantifying the effect that perturbing individual train instances would have on a specific test prediction. 
However, even approximating the IF is computationally expensive, to the degree that may be prohibitive in many cases. 
Might simpler approaches (e.g., retrieving train examples most similar to a given test point) perform comparably? 
%In this work, we investigate the quality of these explanations in comparison with their simpler counterparts, i.e., using similarity metrics.    
%Hindered with the computational complexity of these models, we first show the competence of their simpler approximations.
In this work, we evaluate the degree to which different potential instance attribution agree with respect to the importance of training samples.
We find that simple retrieval methods yield training instances that differ from those identified via gradient-based methods (such as IFs), but that nonetheless exhibit desirable characteristics similar to more complex attribution methods. 
%We have release the open-source implementation of our models at 
Code for all methods and experiments in this paper is available at: \url{https://github.com/successar/instance_attributions_NLP}.
%Adopting a leave-one-out evaluation, we then demonstrate that using similarity metrics over %latent space representation 
%learned representations can identify training points that are similar  %provide explanations 
%similar to gradient based methods. 
%Finally, assessing the quality of the most influential training samples, we observe the competitive quality of the similarity based approaches in comparison to existing attribution methods. 
\end{abstract}
%%%%%%%%%%%%%%%%%%%%%%%%%%%%%
\section{Introduction}
\label{section:interpretability}

Interpretability methods %have been introduced To help users
are intended to help users understand % machine learning 
model predictions \citep{ribeiro2016should, lundberg2017unified,sundararajan2017axiomatic, gilpin2018explaining}. 
%Although mostly these methods provide explanation utilizing the target instance features, the family of instance-based attribution methods introduced to explain models' prediction by computing the influence of each training sample 
In machine learning broadly and NLP specifically, such methods have focused on feature-based explanations that highlight parts of inputs `responsible for' the specific prediction. %, missing explanation on where the model came from.
Feature attribution, however, does not communicate a key basis for model outputs: training data.
%It is, however, unclear why the model came to behave that way.
%\sameer{need a better connective: why feature-based aren't enough}
Recent work has therefore considered %an alternative approach of 
methods for surfacing training examples that were influential for a specific prediction~\citep{koh2017understanding,yeh2018representer,pezeshkpour2019investigating,charpiat2019input,barshan2020relatif,han2020explaining}. 
%Despite efforts to validate the quality of such \emph{instance-attribution} methods, their high computational complexity and unclear usefulness on text data hinder the adoption of these approaches for pretrained language models.
While such \emph{instance-attribution} methods provide an appealing mechanism to identify sources that led to specific predictions (which may reveal potentially problematic training examples), they have not yet been widely adopted, at least in part because even approximating influence functions \cite{koh2017understanding}---arguably the most principled attribution method---can be prohibitively expensive in terms of compute. 
%%%%%%%%%%%%%%%%%%
\begin{figure}
    \centering
    \includegraphics[scale=.4]{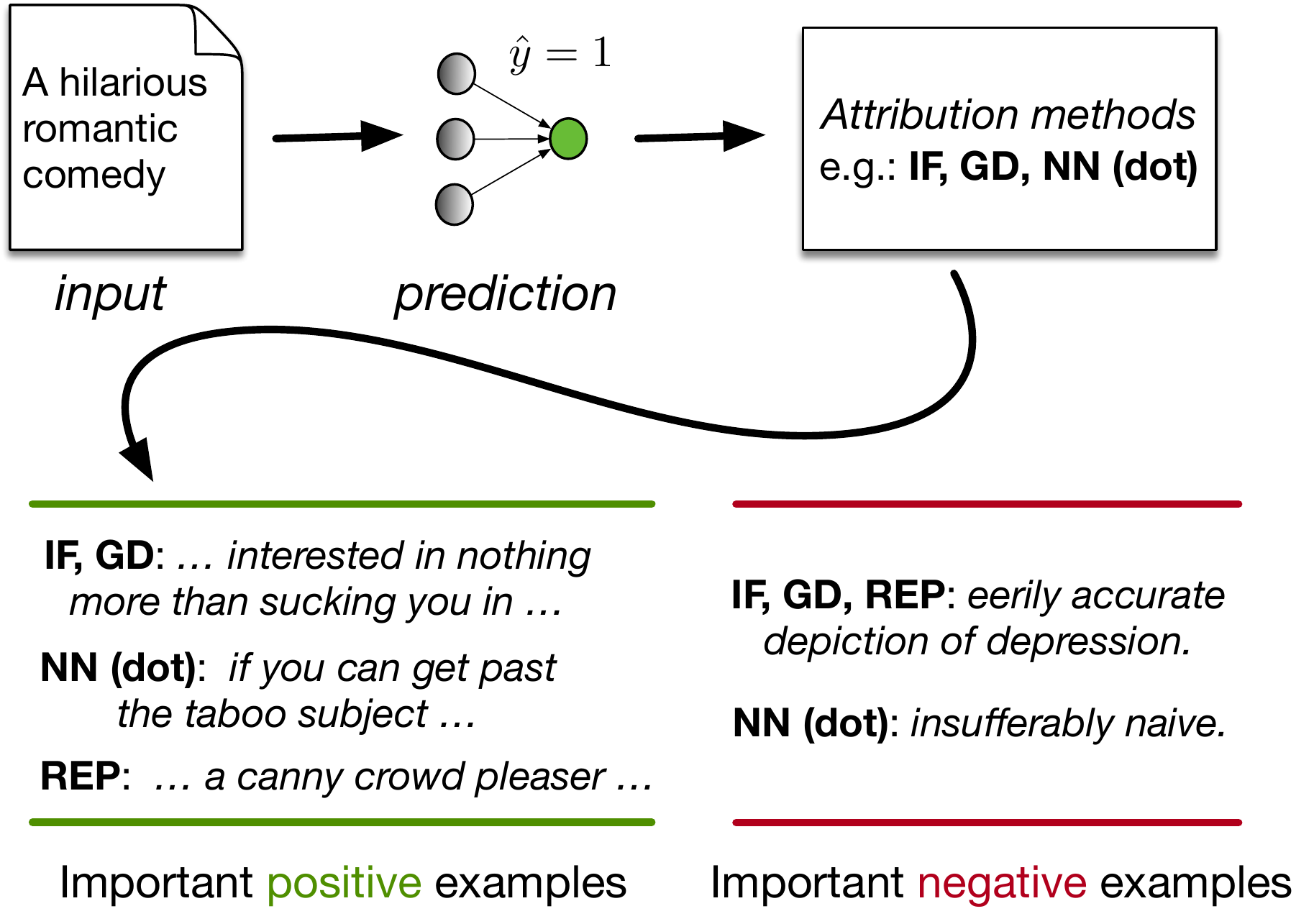}
    \caption{Attribution methods score train examples in terms of their importance to a particular prediction. In this work, we compare several such methods, e.g., Influence Functions (IF) and its variants (GD), Representer Points (REP) and similarity measures (NN).}%\sameer{mention GD and NN}}
    \label{fig:examp}
    \postspace{}
      \minipostspace{}
\end{figure}
%%%%%%%%%%%%%%%%
Is such complexity necessary to identify `important' training points? 
Or do simpler methods (e.g., attribution scores based on similarity measures between train and test instances) yield comparable results?
In this paper, we set out to evaluate and compare instance attribution methods, including relatively simple and efficient approaches \citep{rajani2020explaining} in the context of NLP (Figure \ref{fig:examp}).
We design qualitative evaluations %, we are interested in answering 
intended to probe the following research questions: (1) How correlated are rankings induced by gradient and similarity-based attribution methods (assessing the quality of more efficient approximations)? % \sameer{broken sentence}
(2) What is the quality of explanations in similarity methods compared to gradient-based ones (clarifying the necessity of adopting more complex methods)? 
% We validate the quality of similarity-based methods in comparison to gradient-based ones, shedding light on the necessity of adopting more complex methods. \sameer{phrase (2) as a question}

% bcw: I think we already made these points in the preceding
%Instance-based explanations mostly utilize model's \textit{Gradient} (\textit{Hessian}) to approximate influence of training samples on target instance prediction. 
%Although these methods provide accurate approximations, they suffer from high computational complexity. 
%On the other hand, using similarity metrics on latent space representation to identify influential training samples provides a more efficient alternative, but it is not clear how quality is their explanations.
%These issues are more crucial in pretrained language models; 1) they contain an excessively huge number of parameters and 2) there are not many previous works study the interpretability of these models \citep{}. 

%In this work, tackling instance-based attribution methods on pretrained language models, through 

We evaluate instance-based attribution methods on two datasets: binarized version of the Stanford Sentiment Treebank (SST-2; \citealt{socher2013recursive}) and the Multi-Genre NLI (MNLI) dataset \citep{williams2017broad}.
We investigate the correlation of more complex attribution methods with simpler approximations and variants (with and without use of the Hessian).  %providing an applicable solution for large models such as pretrained language models. 
Comparing explanation quality of gradient-based methods against simple similarity retrieval using leave-one-out \citep{basu2020influence} and randomized-test \citep{hanawa2021evaluation} analyses, we show that simpler methods are fairly competitive. 
Finally, using the HANS dataset \citep{mccoy2019right}, we show the ability of similarity-based methods to surface artifacts in training data.
%  evaluating models behavior in a controlled setting, 

% \sameer{something about no need Hessian? tht also shows we don't need complexity}
 
% % %%%%%%%%%%%%%%%%%%%%
% \begin{figure*}
%     \centering
%     \includegraphics[width=\linewidth]{fig/examp1.png}
%     \caption{Top positive and negative samples based on different attribution methods.}
%     \label{fig:examp}
% \end{figure*}

% % %%%%%%%%%%%%%%%%%%%%%
% %%%%%%%%%%%%%%%%%%%%
%\begin{figure*}
%    \centering
%    \includegraphics[width=\linewidth]{fig/examp2.png}
%    \caption{Top positive and negative samples based on different attribution methods.}
%    \label{fig:examp}
%\end{figure*}

% %%%%%%%%%%%%%%%%%%%%%

%%%%%%%%%%%%%%%%%%%%%%%%%%%%%%
\section{Attribution Methods}
%In this section, we introduce existing instance-based attribution methods, setting up notations and required background for the remaining of the paper. 

\paragraph{Similarity Based Attribution}
Consider a text classification task in which we aim to map inputs $x_i$ to labels $y_i \in Y$. 
We will denote learned representations of $x_i$ by $f_i$ (i.e., the representation from the penultimate network layer). %output of the last layer in our neural network). 
To quantify the importance of training point $x_i$ on the prediction for target sample $x_t$, we calculate the similarity in embedding space induced by the model.\footnote{To be clear, there is no guarantee that similarity reflects `influence' at all, but we are interested in the degree to which this simple strategy identifies `useful' training points, and whether the ranking implied by this method over train points agrees with rankings according to more complex methods.}
To measure similarity we consider three measures: \textit{Euclidean} distance, \textit{Dot} product, and \textit{Cosine} similarity. 
Specifically, we define similarity-based attribution scores as:
\noindent \textbf{NN EUC} = $-\Vert f_t - f_i \Vert ^2 $, \textbf{NN COS} = ${\text{cos}}(f_t, f_i)$, and \textbf{NN DOT} = $\langle f_t, f_i \rangle$.

%\noindent 
To investigate the effect of fine-tuning on these similarity measures, %in addition to the nearest neighbour baselines, 
we also derive rankings based on similarities between untuned sentence-BERT \citep{reimers2019sentence} representations.
%We denote these variants parenthetically, e.g., \textbf{NN-DOT (sent)}. 

\paragraph{Gradient Based Attribution}
\textit{Influence Functions (IFs)} were proposed in the context of neural models by \citet{koh2017understanding} to quantify the contribution made by individual training points on specific test predictions.
Denoting model parameter estimates by $\hat{\theta}$, the IF approximates the effect that upweighting instance $i$ by a small amount---$\epsilon_i$---would have on the parameter estimates 
% \sameer{unsure if ``weighting'' is the right word here} 
(here $H$ is the Hessian of the loss function with respect to our parameters): 
%\begin{align*}
    $\frac{d\hat{\theta}}{d\epsilon_i} = - H_{\hat{\theta}}^{-1} \nabla_{\theta}{\mathcal{L}(x_i, y_i, \hat{\theta})}$.  
%\end{align*}
% \noindent And 
This estimate can in turn be used to derive the effect on a specific test point $x_{\text{test}}$: $\nabla_{\theta}{\mathcal{L}(x_{\text{test}}, y_{\text{test}}, \hat{\theta})^T} \cdot \frac{d\hat{\theta}}{d\epsilon_i}$.
%The use of IF in NLP was recently explored by \citet{han2020explaining}.

%In addition to IF, three other similar variations of gradient-based attributions introduced as:
Aside from IFs, we consider three other similar gradient-based variations: % defined as: % that have been recently proposed:

%\begin{itemize}[leftmargin=*]
%  \vspace{.25em}
% \noindent 
(1)~\text{RIF} = $\text{cos}(H^{-\frac{1}{2}} \gradloss(x_{\text{test}}), H^{-\frac{1}{2}} \gradloss(x_i))$. %introduced in

% \vspace{.25em}
% \noindent
(2)~\text{GD} = $\langle \gradloss(x_{\text{test}}), \gradloss(x_i) \rangle$, and

%\citep{charpiat2019input}.
% \vspace{.25em}
% \noindent 
(3)~\text{GC} = $\text{cos}(\gradloss(x_{\text{test}}), \gradloss(x_i))$. % \citep{charpiat2019input}.
%\end{itemize}
 %  RelatIF (
%  \vspace{.5em}
%\noindent 

RIF was proposed by \citet{barshan2020relatif}, while GD and GC by \citet{charpiat2019input}.

\textit{Representer Points} (REP; \citealt{yeh2018representer}) introduced to
approximate the influence of training points on a test sample by defining a classifier as a combination of a feature extractor and a ($L2$ regularized) linear layer: $\phi(x_i , \theta)$. 
% The objective is to optimize for $\text{argmin} \frac{1}{n}\sum_i^n \loss(x_i, y_i, \theta) + \lambda ||\theta_l||$, where $\lambda$ is a regularization coefficient and $\theta_l$ denote the linear layer parameters.
\citet{yeh2018representer} showed that for such models the output for any target instance $x_t$ can be expressed as a linear decomposition of ``data importance'' of training instances: 
%\begin{align*}
$\phi(x_t , \theta^*) = \sum_i^n \alpha_if_i^\top f_t  = \sum_i^n k(x_t, x_i, \alpha_i)$, 
%\end{align*}
where $\alpha_i = \frac{1}{-2\lambda_n}\frac{\partial  \loss(x_i, y_i, \theta)}{\partial\phi(x_i , \theta)}$. 
% Thus, the $k(x_t, x_i, \alpha_i)$ captures the importance of training sample $x_i$ for the prediction of $x_t$. 
% See \citet{yeh2018representer} for more details.

%%%%%%%%%%%%%%%%%%%%%
\begin{figure*}
\captionsetup[subfigure]{justification=centering}
\begin{subfigure}{.5\linewidth}
    \centering
    \includegraphics[width=\textwidth]{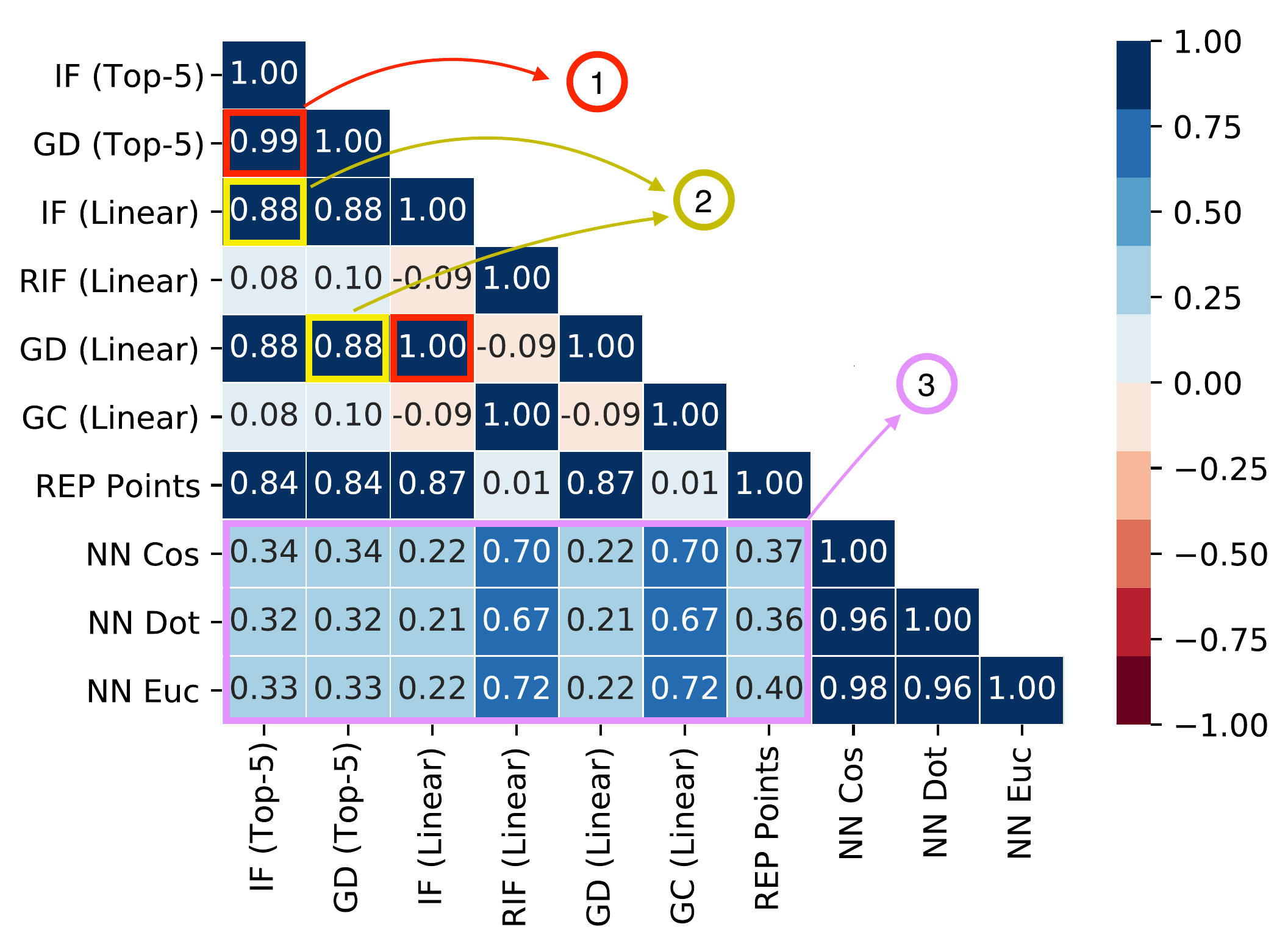}
    \caption{Spearman Correlation on SST.}
    \label{fig:inter-corr-sst}
\end{subfigure}
\begin{subfigure}{.5\linewidth}
    \centering
    \includegraphics[width=\textwidth]{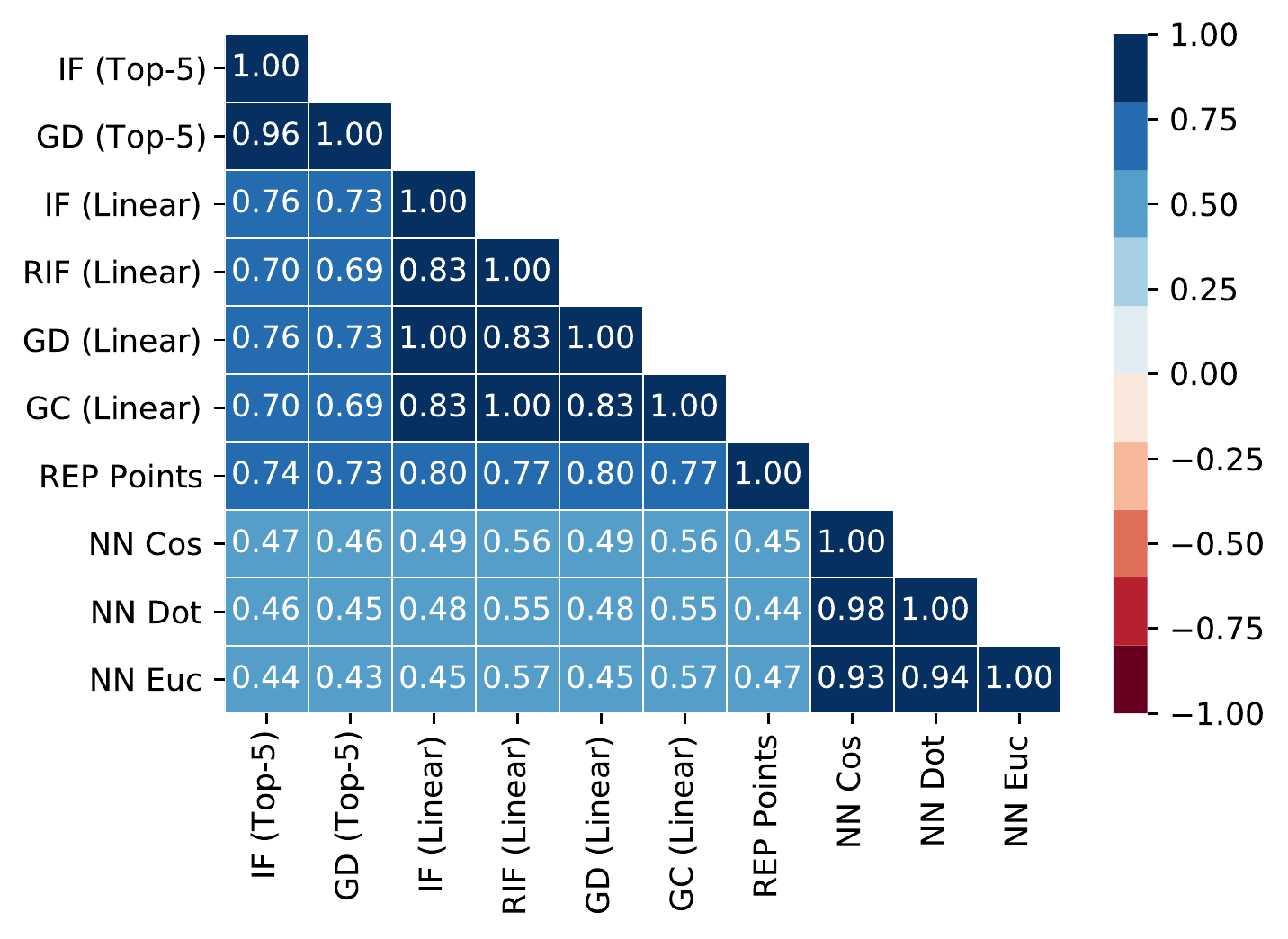}
    \caption{Spearman Correlation on MNLI}
    %between various \\ attribution methods for MNLI.}
    \label{fig:inter-corr-mnli}
\end{subfigure}
    \caption{The similarity between influence of training samples for different pairs of attribution methods on the SST and MNLI datasets was measured via Spearman Correlation. %\sameer{between what? clarify}
    \textcircled{\small 1} = Using Hessian does not change the ordering of training examples. \textcircled{\small 2} = Using more layers of BERT in IF approximation does not much affect the ordering. \textcircled{\small 3} = NN metrics are not well correlated with gradient-based ones.}
    \label{fig:inter-corr}
    \postspace{}
        \minipostspace{}
\end{figure*}
%%%%%%%%%%%%%%%
%%%%%%%%%%%%%%%%%%%%%%%%%%%%%
\section{Experimental Setup}

\paragraph{Datasets}
To evaluate different attribution methods, we conduct several experiments on sentiment analysis and NLI tasks, following prior work investigating the use of IF specifically for NLP \cite{han2020explaining}. 
We adopt a binarized version of the Stanford Sentiment Treebank (SST-2; \citealt{socher2013recursive}), and the Multi-Genre NLI (MNLI) dataset \citep{williams2017broad}. 
For fine-tuning on MNLI, we randomly sample 10k training instances.
Finally, to evaluate the ability of instance attribution methods to reveal annotation \emph{artifacts} in NLI, 
we randomly sampled 1000 instances from the HANS dataset (more details in the Appendix).

\paragraph{Models}
We define models for both tasks on top of BERT \citep{devlin2019bert}, tuning hyperparameters on validation data via grid search. %and performing grid search to find the best hyperparameters.
Our models achieve $90.6 \%$ accuracy on SST and $71.2 \%$ accuracy on MNLI (more details in the Appendix). 
%Let's us note that, it is shown that influence functions can be applied to the BERT model in practice for the tasks we consider in this work \citep{han2020explaining}.

\paragraph{Computing the IF for BERT}%Computing the Influence Function for the Full BERT Parameter Set} %Details Concerning 
Deriving the IF for all parameters $\theta$ of a BERT-based model requires deriving the corresponding Inverse Hessian.
We compute the Inverse Hessian Vector Product (IHVP) $H^{-1}\gradloss(x, y, \theta)$ directly because storing the entire matrix of $|\theta|^2$ elements is practically impossible (requiring $\sim$12 PB of storage). 
We approximate the IHVP using the LiSSa algorithm \cite{lissa}. 
%While less compute-intensive, 
This method is still expensive to run and is sensitive to the norm of the IHVP approximation. %unstable with respect to norm of the IHVP approximation. 
Therefore, for computational reasons we consider IF with respect to the subset of parameters that correspond to the top five layers [IF (Top-5)], and only the last linear layer [IF (linear)], resulting in a few orders of magnitude faster procedure (the algorithm becomes increasingly unstable as we incorporate additional layers). 
We also use a large scaling factor to aid convergence.

%%%%%%%%%%%%%%%
\begin{table*}
\small
\centering
\begin{tabular}{lccccccc}
\toprule
&\multirow{2}{*}{\bf Method}&\multicolumn{2}{c}{\bf avg($\Delta$)-SST}&\multicolumn{2}{c}{\bf avg($\Delta$)-MNLI}&\multicolumn{2}{c}{\bf Spearman}\\
\cmidrule(lr){3-4}
\cmidrule(lr){5-6}
\cmidrule(lr){7-8}
& & Remove-50 & Remove-500& Remove-50 & Remove-500&SST&MNLI \\%\textbf{stddev($\Delta$)}\\
\midrule
&Random (50 runs) & -0.028&-0.021&-0.039&-0.029&-&-\\
% &&std(0.025)&std(0.02)&std(0.035)&std(0.035)&-&-\\

\midrule
% \multirow{3}{*}{\rotatebox[origin=c]{90}{\bf Sim}} &
\multirow{3}{*}{\bf Similarity} &
NN EUC & -0.028&\bf -0.540&-0.102&-0.266&\color{green!50!black!50}{0.056}&\color{green!50!black!70}{0.023}\\
&NN COS&  -0.072&-0.430&-0.088&-0.306&\color{green!50!black!50}{0.045}&\color{green!50!black!80}{0.018}\\
&NN DOT&  -0.059&-0.513&\bf -0.106&-0.273&\color{green!50!black}{ 0.005}&\color{green!50!black}{ -0.002}\\

\midrule

% \multirow{5}{*}{\rotatebox[origin=c]{90}{\bf Gradient}} &
\multirow{5}{*}{\bf Gradient} &
IF &  -0.054&-0.526&-0.042&-0.407&\color{red!80!black!50}{-0.296}&\color{green!50!black!80}{0.018}\\
&REP & \bf -0.114&-0.490&-0.002&-0.230&\color{red!80!black!50}{-0.217}&\color{green!50!black!50}{0.053}\\
&RIF& -0.071& -0.537&-0.068&-0.347&\color{green!50!black!70}{-0.021}&\color{green!50!black!80}{0.013}\\
&GD&-0.058&-0.516&-0.022&\bf -0.446&\color{red!80!black!50}{-0.290}&\color{green!50!black!80}{0.017}\\
&GC&-0.082&-0.528&-0.030&-0.279&\color{green!50!black!70}{-0.021}&\color{green!50!black!80}{0.012}\\
\bottomrule
\end{tabular}
\caption{Average difference ($\Delta$) between predictions made after training on (i) all data and (ii) a subset in which we remove the top-50/top-500 most important training points, according to different methods (Random on both of the benchmarks has standard deviation around $0.02$). We also report the Spearman correlation between the ranking induced by each approach using a trained model and the same ranking when a randomly initialized model is used.}%$K$ is the \#instances removed, $N$ is the number of test points, $P$ is the number of repetitions of random, $Q$ number of times you retrain for each testpoint+attribution method.}
\label{tab:remove-acc-SA}
    \postspace{}
        \minipostspace{}
\end{table*}
%Studying the effect of different attribution method by
%%%%%%%%%%%%%%%%%%%%
%%%%%%%%%%%%%%%%%%%%%%%%%%%%%%
\section{Experiments}
In this section, we first investigate the correlation between different methods. 
Then, to study the quality of explanations we conduct leave-some-out experiments, and further analyze attribution methods on HANS data.
We consider four evaluations (more analyses and experimental details in the Appendix). 
% \sameer{maybe the following can be in corresponding sections only, just a high level summary sentence should be enough here}

\vspace{.2em}
\noindent(1) Calculating the \textit{correlation} of each pair of attribution methods, 
%' ranking for a subset of train and test data, to 
assessing whether simple methods induce rankings similar to more complex ones. %showing the quality of more efficient approximations. 

\vspace{.2em}
\noindent(2) \textit{Removing the most influential samples} according to each method, retrain, and then observe the change in the predicted probability for the originally predicted class, with the assumption that more accurate attribution methods will cause more drop. 

\vspace{.2em}
\noindent(3) We follow \textit{randomized-test} from \cite{hanawa2021evaluation} and measure the ranking correlation of methods for (a) randomly initialized and (b) trained models, under the assumption that high correlation here would suggest less meaningful attribution.
% Adopting randomize-test from \citet{hanawa2021evaluation}, conjecturing that a better attribution method demonstrates less correlation between ranking give for a random model and a well-trained one. As our random model, we consider our model without any fine-tuning. 

% \vspace{.25em}
\vspace{.2em}
\noindent(4) We measure the degree to which the methods recover examples that exhibit \textit{lexical overlap} when tested on the HANS dataset \cite{mccoy2019right}. 
This extends a prior analysis of IF \cite{han2020explaining}, considering alternative attribution methods. %conjecturing that better attribution methods will give higher influence to samples that have a high rate of overlap when the model predicts the test target utilizing this artifact

%%%%%%%%%%%%%%%
\paragraph{Attribution Methods' Correlation}
%To compare the quality of more complex attribution methods to simpler approaches, i.e., only considering the linear classification layer, and further evaluate similarity-based methods, w
We calculate the Spearman correlation between scores assigned to training samples by different methods, allowing us to compare their similarities.
More specifically, we randomly sample 100 test and 500 training samples from datasets and calculate the average resultant Spearman correlations.
%The correlation between different attribution methods is depicted 

%\paragraph{SST:} 
We report attribution methods' correlation on SST and MNLI datasets in Figure \ref{fig:inter-corr} (a more complete version of these figures is in the Appendix). 
We make the following observations. 
(1) Gradient methods w/wo normalization appear similar to each other, e.g., GC is similar to RIF and IF is similar to GD, suggesting that Hessian information may not be necessary to provide meaningful attributions (GD and GC do not use the Hessian). 
(2) There is a high correlation between IF calculated over the top five layers of BERT and IF over only the last linear layer.
(3) There is only a modest correlation between similarity-based rankings and gradient-based methods, suggesting that these do differ in terms of the importance they assign to training instances. 
We report a proportion of common top examples between IF (Top-5) and IF (Linear) in the Appendix, providing further evidence of the high correlation between these methods.

% \paragraph{MNLI:} Similar study on correlation of attribution methods for MNLI data is provided in Figure \ref{fig:inter-corr-mnli}.  
% In addition to previous observations, we can see a higher degree of correlation between gradient-based attribution (except for IF top-5).
% Moreover, similarity-based methods appear to have less correlation on MNLI in comparison to the SST dataset.

% %%%%%%%%%%%%%%%
% \begin{table}
% \small
% \centering
% \begin{tabular}{cccc}
% \toprule
% &\multirow{2}{*}{\bf Method}&\multicolumn{2}{c}{\bf Lexical Overlap Rate}\\
% \cmidrule(lr){3-4}
% & & top-1 & top-10 \\%\textbf{stddev($\Delta$)}\\
% \midrule
% & Random &0.40&0.40\\
% % \midrule
% % \multirow{3}{*}{{\bf Sen-Bert}} &
% % NN-EUC & 0.39 & 0.41\\
% % &NN-COS&  0.38& 0.39\\
% % &NN-DOT& 0.39 & 0.40\\
% \midrule
% \multirow{3}{*}{{\bf Similarity}} &
% NN EUC & \bf 0.56 & \bf 0.57\\
% &NN COS&  \bf 0.56 & 0.56\\
% &NN DOT& 0.44 & 0.44\\

% \midrule

% \multirow{5}{*}{{\bf Gradient}} &
% IF &  0.43 & 0.44\\
% &REP & 0.43 & 0.35\\
% &RIF& 0.54 & 0.55\\
% &GD& 0.43 & 0.44\\
% &GC& 0.55 & 0.56\\
% \bottomrule
% \end{tabular}
% \caption{Average lexical overlap rate  between premise and hypothesis in top-$k$ most influential samples for test instances mispredicted as entailment.}
% \label{tab:HANS}
%     % \postspace{}
%     %     \minipostspace{}
% \end{table}
% %%%%%%%%%%%%%%%%%%%%
%%%%%%%%%%%%%%%
\begin{table}
\small
\centering
\begin{tabular}{lrrr}
&\multirow{2}{*}{\bf Method}&\multicolumn{2}{c}{\bf Lexical Overlap Rate}\\
\cmidrule(lr){3-4}
& & top-1 & top-10 \\%\textbf{stddev($\Delta$)}\\
\midrule%\hline
& Random &0.40&0.40\\
\midrule
\multirow{3}{*}{{\bf Sen-Bert}} &
NN EUC & 0.39 & 0.41\\
&NN COS&  0.38& 0.39\\
&NN DOT& 0.39 & 0.40\\
\midrule
\multirow{3}{*}{{\bf Sim}} &
NN EUC & \bf 0.56 & \bf 0.57\\
&NN COS&  \bf 0.56 & 0.56\\
&NN DOT& 0.44 & 0.44\\

\midrule

\multirow{5}{*}{{\bf Gradient}} &
IF &  0.43 & 0.44\\
&REP & 0.43 & 0.35\\
&RIF& 0.55 & 0.56\\
&GD& 0.43 & 0.44\\
&GC& 0.55 & 0.56\\
\hline
\end{tabular}
\caption{Average lexical overlap rate  between premise and hypothesis in top-$k$ most influential samples for test instances mispredicted as entailment.}
\label{tab:HANS}
    \postspace{}
    % \postspace{}
          \minipostspace{}
\end{table}
%%%%%%%%%%%%%%%%%%%%

%%%%%%%%%%%%%%%%%%%
\paragraph{Removing `Important' Samples} %on the Probability of Predicted Class:}
In Table \ref{tab:remove-acc-SA} we report the average results of removing the top-$k$ most important training samples for 50 random test samples using different attribution methods. 
We only consider the linear version of methods in the remainder of the paper. 
All methods seem effective, compared to random sampling. 
Perhaps surprisingly, for both tasks at least one of the similarity-based approaches performs comparably or better than gradient-based methods, in the sense that removing the top examples according to similarity yields reductions in the predicted probability (which is what one would intuitively hope).
Finally, it seems that the models applying some form of normalization to the gradient (i.e., RIF and GC) perform more consistently.
This is consistent with contemporaneous work of 
% \sameer{say contemporaneous} 
\citet{hanawa2021evaluation} which argues that this is a consequence of large gradient magnitudes for some samples dominating when normalization is not used. %in the case of attribution methods that are not using normalization.
Upon investigating high influential training samples, we observed that similarity-based approaches seem to yield more diverse “top” instances compared to gradient-based ones. We also found that normalization in gradient-based methods made a large difference. Generic IF-based ranking tends to be dominated by high loss training examples across test examples, whereas normalization provides more diverse top training examples. 
Further, proportions of shared top examples between methods is provided in the Appendix, clarifying their similar performance.

%%%%%%%%%%%%%%%%%%%%%%%%%%%%%%%%%%%%%%%%
\paragraph{Randomized-Test}
We report the Spearman correlation between trained and random models for SST and MNLI data in Table \ref{tab:remove-acc-SA}.
This would ideally be small in magnitude (non-zero values indicate correlation). 
Curiously, gradient-based methods (IF, REP, GD) exhibit negative correlations on the SST dataset.
Overall, these results suggest that gradient-based approaches without gradient normalization may be inferior to alternative methods.
The simple NN-DOT method provides the `best' performance according to this metric.
%, having the similarity metric based on dot product with the best performance.  
% \sameer{tehre is some work (I can find it) that shows that some saliency explanations mostly capture the input, not the conditional distribution. this might be happening here with more complex methods}

%%%%%%%%%%%%%%%%%%%%%%%%%%%%%%%%%%%%%%%
\paragraph{Artifacts and Attribution Methods}
To investigate whether attribution methods can correctly identify training samples with specific artifacts responsible for model predictions we follow \citet{han2020explaining}: This entails randomly choosing 10k samples from MNLI and treating \emph{neutral} and \emph{contradiction} as a single \emph{non-entailment} label for model fine-tuning. 
More specifically, we are interested in target samples that the model mispredicts as \emph{entailment} because of the lexical overlap artifact (lexical overlap is an artifactual indicator of entailment; \citealt{mccoy2019right}). 

The average lexical overlap rate for 1000 random samples from the HANS dataset is provided in Table \ref{tab:HANS}. As a baseline, we also apply similarity-based methods on top of sentence-BERT embeddings, which as expected appear very similar to random correlation. 
One can observe that similarity-based approaches tend to surface instances with higher lexical overlap, compared to gradient-based instance attribution methods. 
Moreover, gradient-based methods without normalization (IF, GD, and REP) perform similar to selecting samples randomly and based on sentence-BERT representations, suggesting an inability to usefully identify lexical overlap. 

% To further investigate the quality of explanations, we conjecture that if a data point be very similar to a training sample, a good method should provide that training sample as the most influential instance. 
% The result of this evaluation is provided in the Appendix further demonstrating the superiority of similarity-based methods.
% \sameer{this feels weird to me.. cutting might be better}

%%%%%%%%%%%%%%%%%%%%%%%%%%%%%%%%%%%%%%%
\paragraph{Computational Complexity}
The computational complexity of IF-based instance attribution methods constitutes an important practical barrier to their use. 
This complexity depends on the number of model parameters taken into consideration. 
As a result, computing IF is effectively infeasible if we consider \emph{all} model parameters for modern, medium-to-large models such as BERT. 

If we only consider the parameters of the last linear layer---comprising $O(p)$ parameters---to approximate the IF, the computational bottleneck will be the inverse Hessian which can be approximated with high accuracy in $O(p^2)$. 
There are ways to approximate the inverse Hessian more efficiently \citep{pearlmutter1994fast}, though this results in worse performance. Similarity-based measures, on the other hand, can be calculated in $O(p)$. 

%We agree that these details are important and will add them along with a computational complexity analysis of different methods as an additional section in a revised version of the paper.

With respect to wall-clock running time, calculating the influence of a single test sample with respect to the parameters comprising the top-5 layers of a BERT-based model for SST classification running on a reasonably modern GPU\footnote{Maxwell Titan GPU (2015).} requires $\sim$5 minutes. 
If we consider the linear variant, this falls to $< 0.01$ seconds. 
Finally, similarity-based approaches require $< 0.0001$ seconds. 
Extrapolating these numbers, it requires about 6 days to calculate IF (top-5 Layer) for all 1821 test samples in SST, while it takes only around 0.2 seconds for similarity-based methods.

%has following timing characteristics on 1 Maxwell Titan GPU (2015):

%IF (Top-5 Layer) ~ 5 min / test example
%IF/RIF/GC/GD (Linear) ~ 0.007 sec / test example
%Similarity ~ <0.0001 sec / test example

%We will add these details in the paper.

%%%%%%%%%%%%%%%%%%%%%%%%%%%%%%%%%%%%%%%%
\section{Conclusions}
% \vspace{-.5em}

Instance attribution methods constitute a promising approach to better understanding how modern NLP models come to make the predictions that they do \cite{han2020explaining,koh2017understanding}.  
However, approximating IF to quantify the importance of train samples is prohibitively expensive. 
In this work, we investigated whether alternative, simpler and more efficient methods provide similar instance attribution scores.

%applying instance-based attribution methods to pretrained language models (such as BERT), we study the validity and quality of simpler more applicable approaches. 
We demonstrated high correlation between (1) gradient-based methods that consider more parameters [IF and GD (top-5)] and their simpler counterparts [IF and GD (linear)], and (2) methods without Hessian information, i.e., IF vs GD and RIF vs GC. 
We considered even simpler, similarity-based approaches and compared the importance rankings over training instances induced by these to rankings under gradient-based methods.
Through leave-some-out, randomized-test, and artifact detection experiments, we demonstrated that these simple similarity-based methods are surprisingly competitive.
This suggests future directions for work on fast and useful instance attribution methods. 
All code necessary to reproduce the results reported in this paper is available at: \url{https://github.com/successar/instance_attributions_NLP}.
%Then, we take a further step to study even more efficient methods by comparing the quality of similarity-based approaches and gradient-based ones through leave-some-out, randomized-test, and artifacts detection (HANS data), demonstrating the competitive quality of studied similarity-based methods.  

%In future work, we intend to extend our evaluations to other existing pretrained language models. 
%Moreover, to better shed light on the connection between active parameters during training and calculating attribution methods, we will study simpler and more computationally efficient models by directly adopting transformers as our classifier. 

% \clearpage

%%%%%%%%%%%%%%%%%%%%%%%%%%%%%%%%%%%%%%
%\section{Broader Impact Statement}
%Pretrained language models are being adopted in an excessively growing number of real-world applications.
\section{Ethical Considerations}
Deep neural models have come to dominate research in NLP, and increasingly are deployed in the real world.
A problem with such techniques is that they are opaque; it is not easy to know why models make specific predictions.
Consequently, modern models may make predictions on the basis of attributes we would rather they not (e.g., demographic categories or `artifacts' in data).

Instance attribution---identifying training samples that influenced a given prediction---provides a mechanism that might be used to counter these issues.
However, the computational expense of existing techniques hinders their adoption in practice. % of such approaches.
By contrasting these complex approaches against simpler alternative methods for instance attribution, we contribute to a better understanding and characterization of the tradeoffs in instance attribution techniques. %, and hope to lead to  
% our findings may lead to better characterization of these approaches, and greater uptake in practice.
This may, in turn, improve the robustness of models in practice, and potentially reduce implicit biases in their predictions.
%An Achilles' heel for wide adoption of these models to different tasks, is the shortage of efficient and accurate interpretability methods, specifically instance-based attributions, to understand and debug them. 
%Although existing instance-based attribution approaches provide tremendous success in providing explanations for a variety of tasks, because of the very high computational complexity of them on pretrained language models, and unclear usefulness of them for these models, they are not applicable for practice.       
%In this work, we try to tackle this problem, by excessively evaluating a variety of instance attribution methods (including relatively simple and efficient approaches) in the context of NLP.
%More specifically, we design qualitative evaluations to assess the quality of simpler and more efficient approaches, providing an applicable solution for pretrained language model. 
 
%In the past few years, with the growth of global concerns regarding climate change and other environmental threats, researchers in AI turns to design models that have efficient computational complexity without affecting the performance. In this work, by asserting the quality of more efficient instance-based attribution methods, we attempt to reduce energy use while providing high-quality explanations for huge pretrained language models. 
%%%%%%%%%%%%%%%%%%%%%%%%%%%%%%%%%%%%
\section*{Acknowledgements}
We thank the anonymous NAACL reviewers for their detailed feedback. 
Further, we also thank Matt Gardner, Sanjay Subramanian, and Dylan Slack for their useful comments. 
This work was sponsored in part by the Army Research Office (W911NF1810328), and funded in part by the NSF grants \#IIS-1756023 and \#IIS-2008956.
The views expressed are those of the authors and do not reflect the policy of the funding agencies.

%%%%%%%%%%%%%%%%%%%%%%%%%%%%%%%%%%%%%
\bibliography{main}
\bibliographystyle{acl_natbib}

% \appendix

% \section{Example Appendix}
% \label{sec:appendix}

% This is an appendix.

% %%%%%%%%%%%%%%%%%%%
\clearpage
\appendix

\section{Experimental Details}
\paragraph{Datasets}
To evaluate different attribution methods, we conduct several experiments on sentiment analysis and NLI tasks, following prior work investigating the use of influence functions specifically for NLP \cite{han2020explaining}. 
We adopt a binarized version of the Stanford Sentiment Treebank (SST-2) \citep{socher2013recursive}, consisting of 6920 training samples and 1821 test samples. 
As our NLI benchmark, we use the Multi-Genre NLI (MNLI) dataset \citep{williams2017broad}, which contains 393k pairs of premise and hypothesis from 10 different genres. 
For model fine-tuning, we randomly sample 10k training instances.
To evaluate the utility of different instance attribution methods in helping to unearth annotation artifacts in NLI, %to interpret attribution methods in a controlled manner on NLI task, 
we use the HANS dataset \citep{mccoy2019right}, which comprises examples exhibiting previously identified NLI artifacts such as lexical overlap between hypotheses and premises.%where hypotheses entail premises with certain artifacts such as lexical overlap. 
We randomly sampled 1000 instances from this benchmark as test data to analyze the behavior of different attribution methods.

\paragraph{Models} As discussed in the paper, we define models for both tasks on top of BERT, tuning hyperparameters on validation data via grid search.
These hyperparameters include the regularization parameter $\lambda=[10^{-1}, 10^{-2}, 10^{-3}]$; learning rate $\alpha=[2\times 10^{-3}, 2\times 10^{-4}, 2\times 10^{-5}, 2\times 10^{-6}]$; number of epochs $\in \{3,7,10,15\}$; and the batch size $\in \{8, 16\}$. 
Our final models achieve $90.6 \%$ accuracy on SST and $71.2 \%$ accuracy on MNLI

%%%%%%%%%%%%%%%%%%%%%%%%%%%%%%%%%%%%%%
\section{Attribution Methods' Correlation}
\begin{figure*}
\begin{subfigure}{.5\linewidth}
    \centering
    \includegraphics[width=\textwidth]{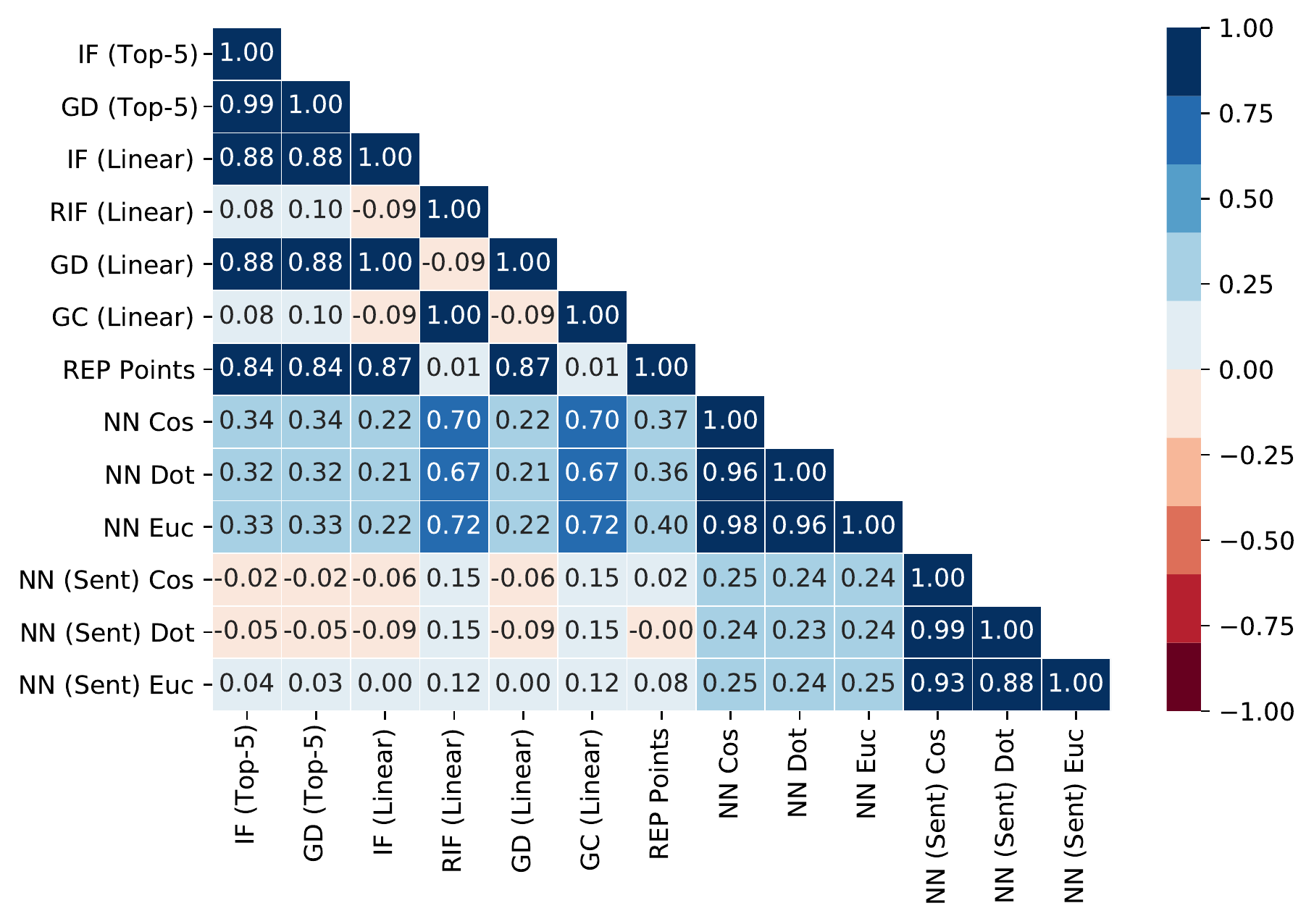}
    \caption{SST.}
\end{subfigure}
\begin{subfigure}{.5\linewidth}
    \centering
    \includegraphics[width=\textwidth]{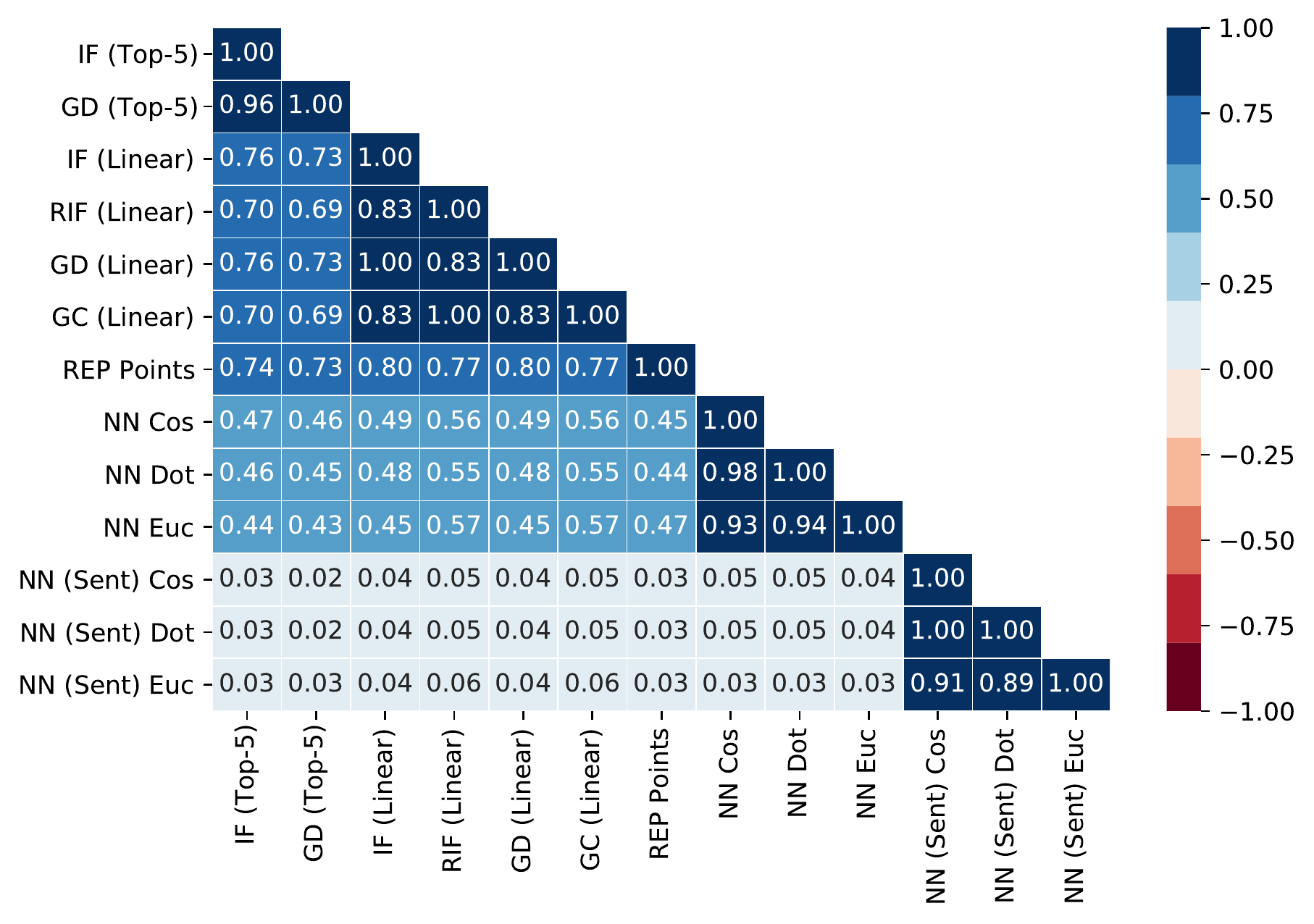}
    \caption{MNLI.}
\end{subfigure}
    \caption{Complete version of correlation matrices.}
    \label{fig:inter-corr-comp}
\end{figure*}

The complete version of Spearman correlation between attribution methods (containing the sentence-BERT) is provided in Figure \ref{fig:inter-corr-comp}. 
As expected, similarity-based approaches based on sentence-BERT show a very small correlation with other methods. 

\begin{figure*}
\begin{subfigure}{.5\linewidth}
    \centering
    \includegraphics[width=\textwidth]{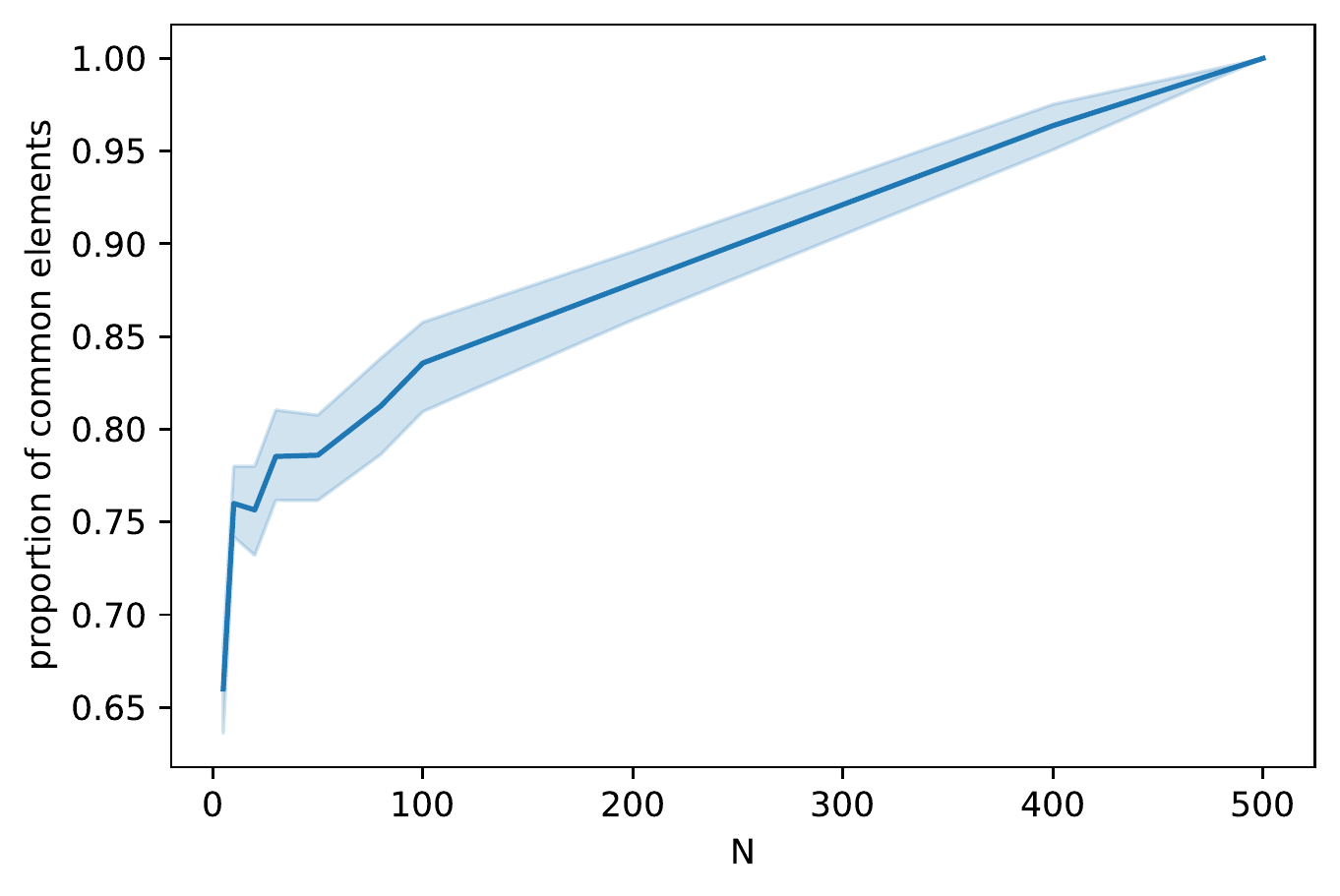}
    \caption{SST.}
\end{subfigure}
\begin{subfigure}{.5\linewidth}
    \centering
    \includegraphics[width=\textwidth]{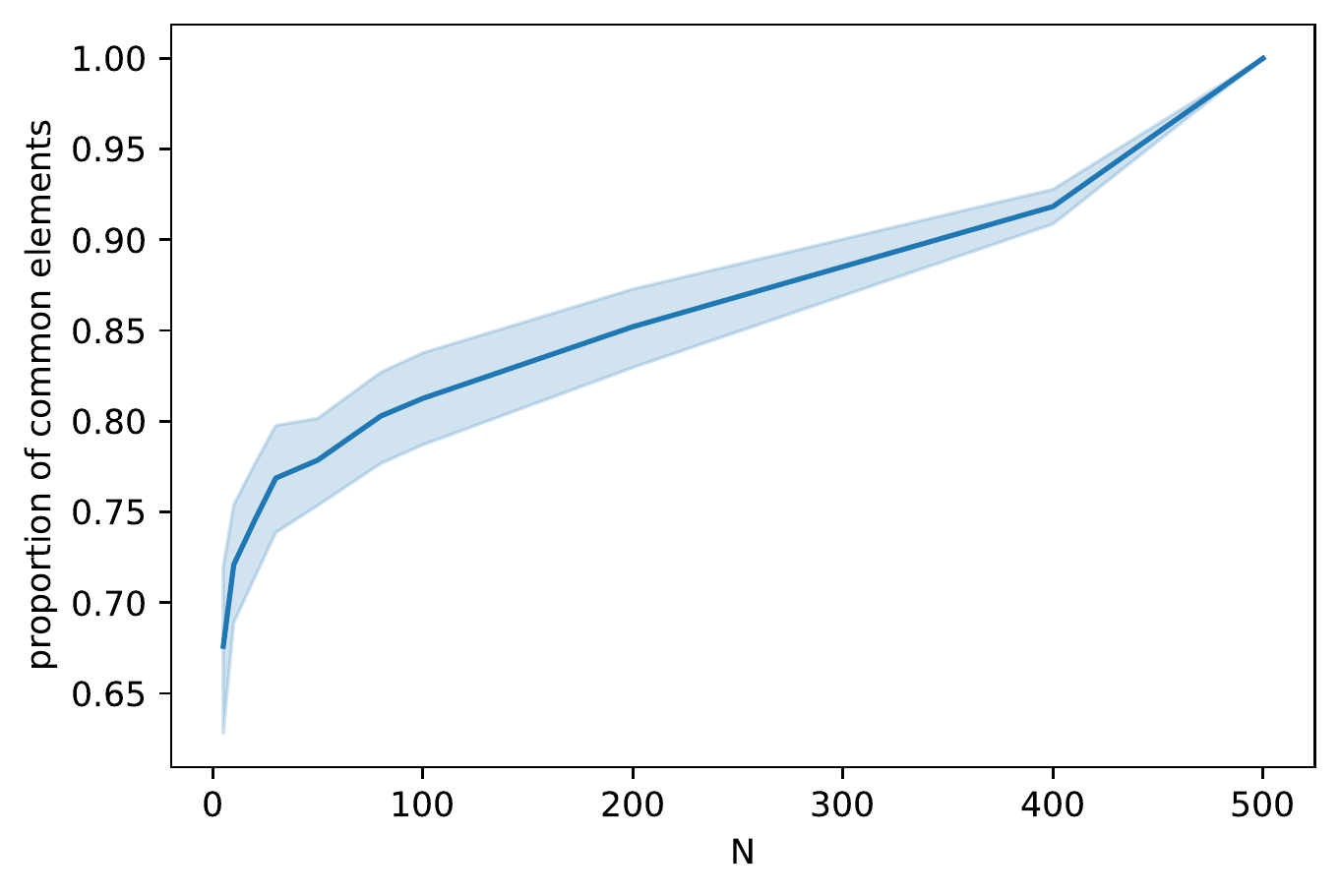}
    \caption{MNLI.}
\end{subfigure}
    \caption{Proportion of common top examples between IF (Top-5) and IF (Linear) Methods. We selected 100 test examples and 500 training examples to compute the attributions over.}
    \label{fig:inter-com}
\end{figure*}

We also provide the proportion of shared examples in the top samples retrieved by IF (top-5) and IF (linear) in Figure \ref{fig:inter-com}. 
One can see that there is a very high correlation between these methods in top samples, validating the high quality of simpler version of IF (IF (linear)) in comparison to the more complex method (IF (top-5)). 

%%%%%%%%%%%%%%%%%%%%%

%%%%%%%%%%%%%%%
%%%%%%%%%%%%%%%%%%%%%

%%%%%%%%%%%%%%%

\section{Removing `Important' Samples}
\begin{figure*}
\begin{subfigure}{.5\linewidth}
    \centering
    \includegraphics[width=\textwidth]{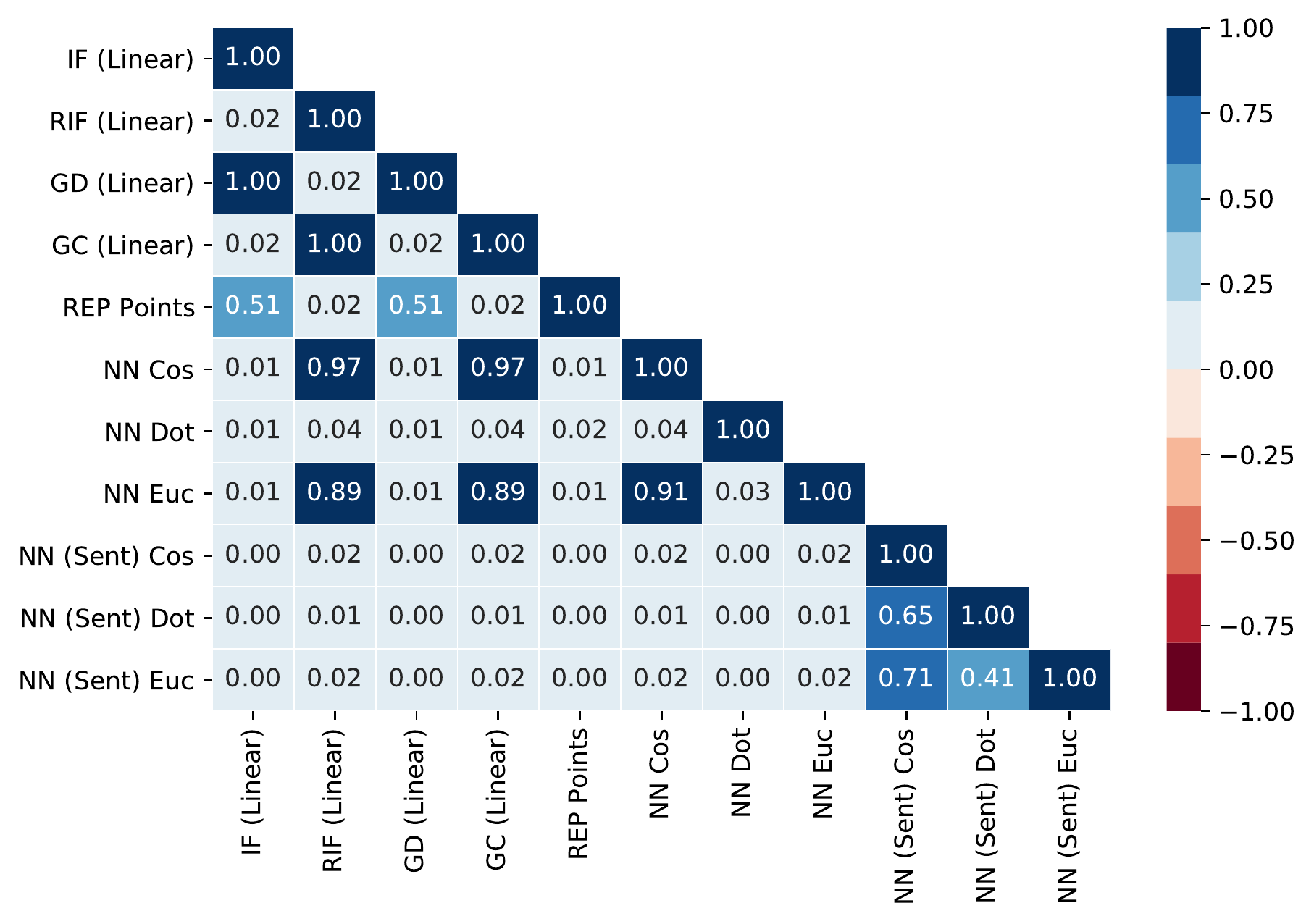}
    \caption{Top-10 in SST.}
\end{subfigure}
\begin{subfigure}{.5\linewidth}
    \centering
    \includegraphics[width=\textwidth]{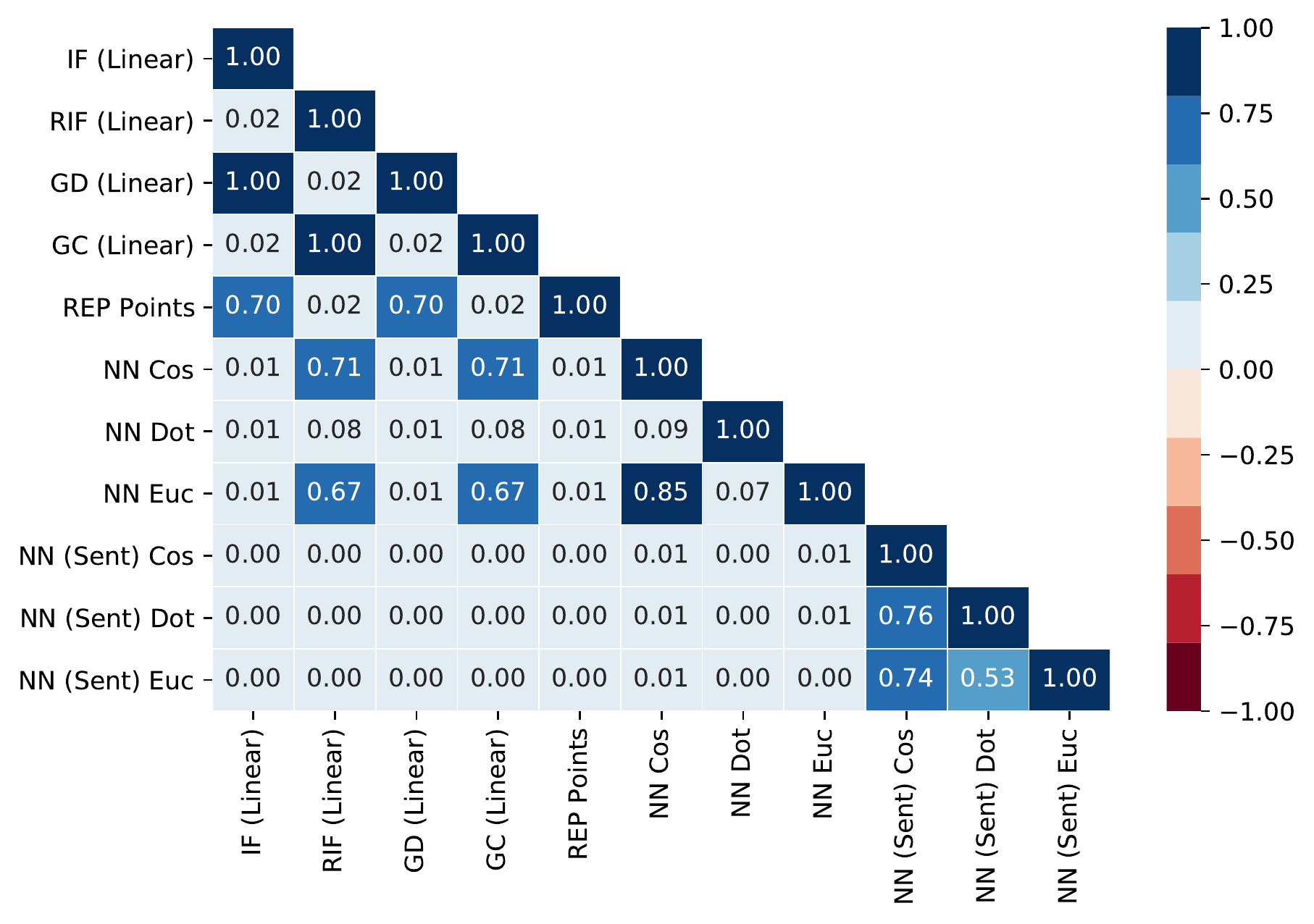}
    \caption{Top-10 in MNLI.}
\end{subfigure}
    \caption{Proportion of common examples in top 10 samples between pairs of attribution methods.}
    \label{fig:inter-com-top10}
\end{figure*}

\begin{figure*}
\begin{subfigure}{.5\linewidth}
    \centering
    \includegraphics[width=\textwidth]{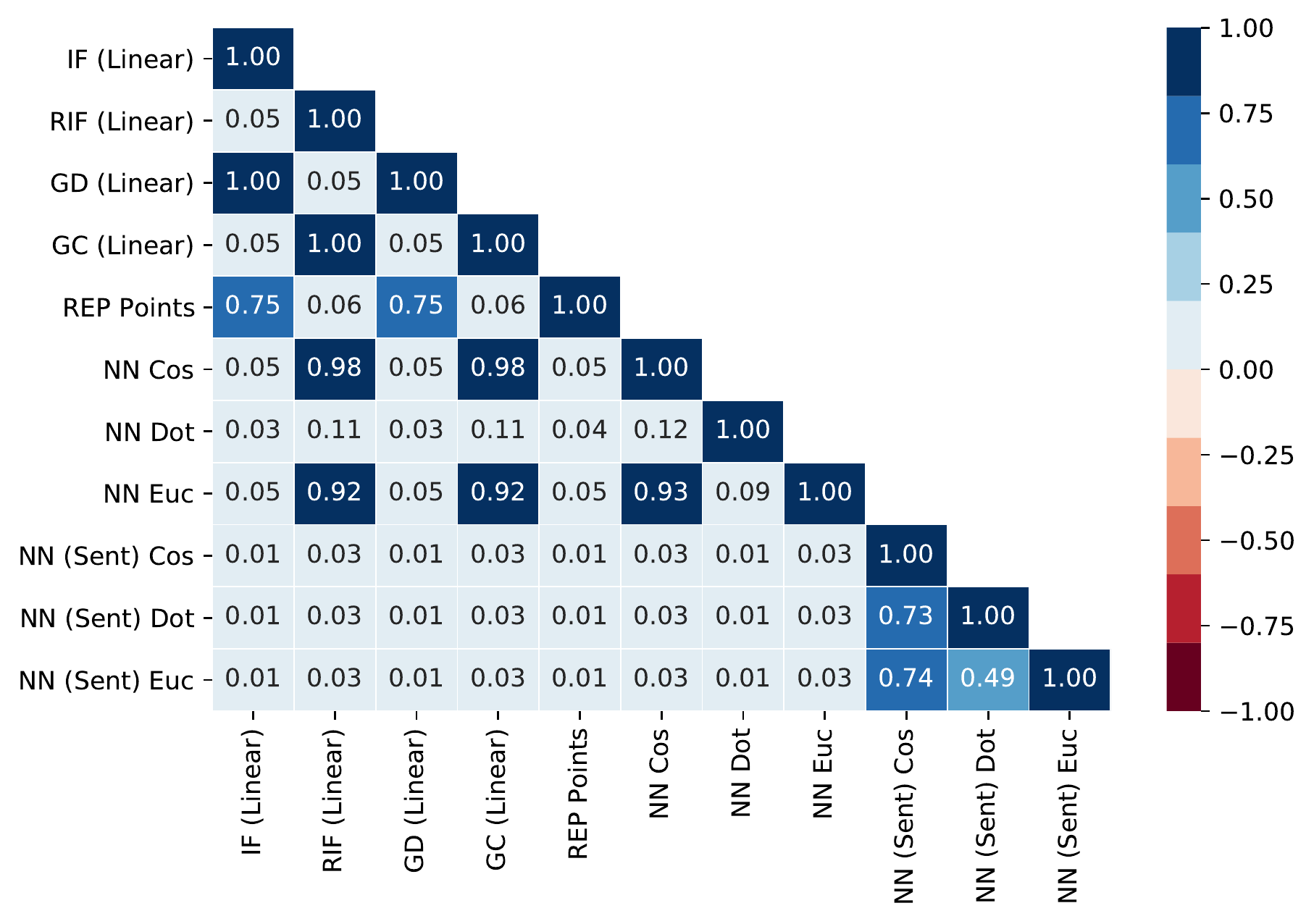}
    \caption{Top-50 in SST.}
\end{subfigure}
\begin{subfigure}{.5\linewidth}
    \centering
    \includegraphics[width=\textwidth]{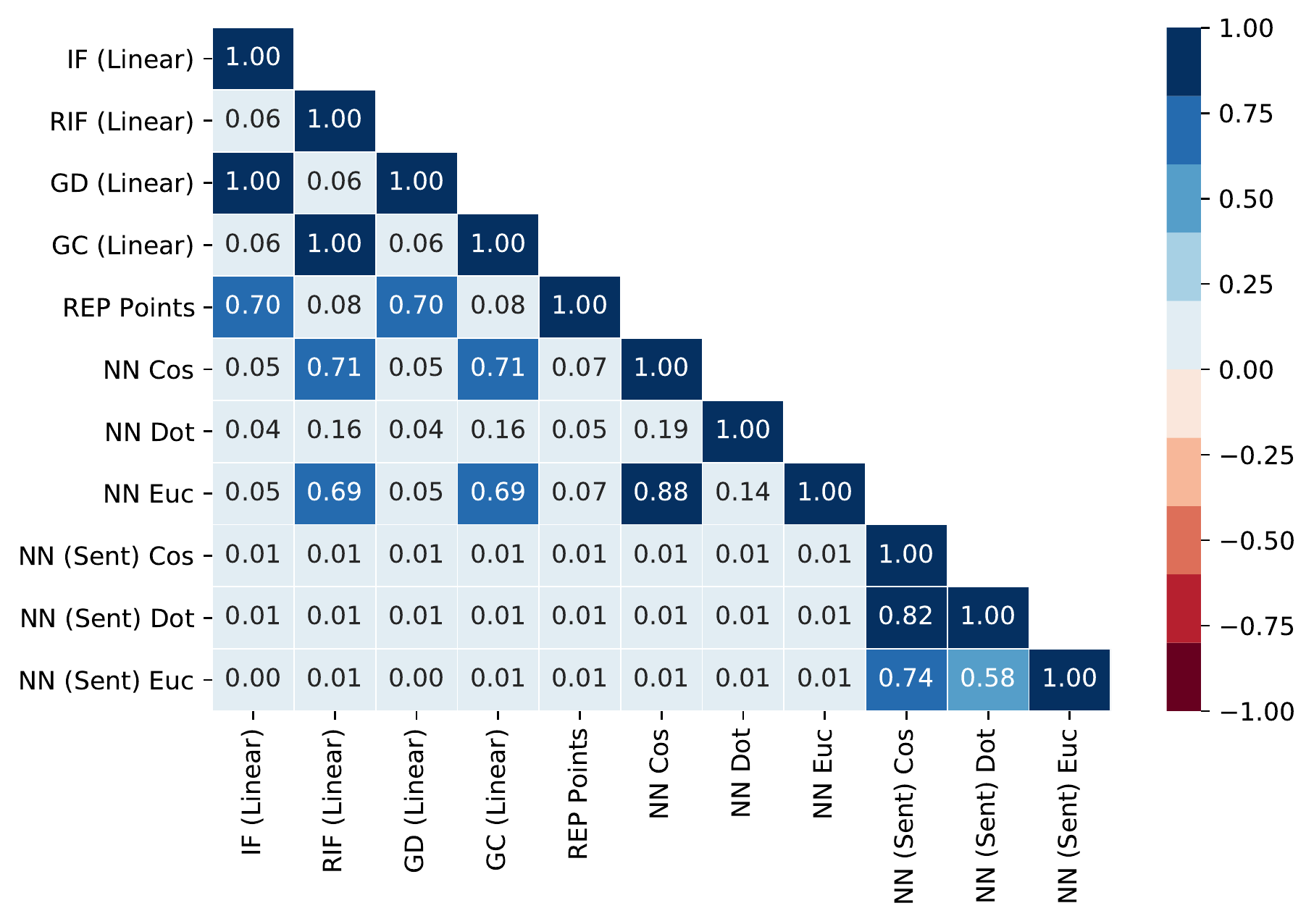}
    \caption{Top-50 in MNLI.}
\end{subfigure}
    \caption{Proportion of common examples in top 50 samples between pairs of attribution methods.}
    \label{fig:inter-com-top50}
\end{figure*}

In this experiment, we first select 50 random test samples (for both MNLI and SST). Then, for each one of these instances, we separately remove top-k (we consider k = 50 and 500) training instances for that test sample, retrain the model, and calculate the change in the model’s prediction for that sample. We report the average changed over the prediction of the selected 50 random test samples in Table~\ref{tab:remove-acc-SA}.
Moreover, the proportion of common examples in top samples between pairs of attribution methods is depicted in Figures \ref{fig:inter-com-top10} and \ref{fig:inter-com-top50}. 
The very high rate between IF vs GD, RIF vs GC, and NN-EUC vs NN-COS pairs, clarify the reason behind the similar performance of these pairs of methods in leave-some-out experiments. 
%%%%%%%%%%%%%%%%%%%%%

%%%%%%%%%%%%%%%
%%%%%%%%%%%%%%%%%%%%%

\section{Near Training Samples Explanations}
To further investigate the quality of the most influential sample based on different attribution methods, we conjecture that a data point very similar to a training sample should recover that sample as the most influential instance. 
We consider four scenarios to create target points similar to training data: (1) using training samples themselves as the target instances for attribution methods; (2) adding a random token to a random place in each training samples; (3) randomly removing a token from each training samples, and; (4) replacing a random token in each training samples with a random token from the dictionary of tokens. 
In the MNLI dataset, we apply each modification to both the premise and hypothesis in each training sample.

The result of this analysis is provided in Tables \ref{tab:mod-SA} and \ref{tab:mod-MNLI}.
We observe that similarity-based methods demonstrate a greater ability to recover the original training samples corresponding to the different targets. 
Moreover, the very low performance of IF, GC, and REP methods is due to the fact that there are training points with high magnitude gradient, which these methods choose as top instances for \emph{any} target sample.

%%%%%%%%%%%%%%%
\begin{table*}
\centering
\begin{tabular}{lrrrrrrrrrr}
\toprule
&\multirow{2}{*}{\bf {Method}}&\multicolumn{2}{c}{\bf Train}&\multicolumn{2}{c}{\bf ADD}&\multicolumn{2}{c}{\bf Remove}&\multicolumn{2}{c}{\bf Replace}\\
\cmidrule(lr){3-4}
\cmidrule(lr){5-6}
\cmidrule(lr){7-8}
\cmidrule(lr){9-10}

& &HIT@1 & HIT@10& HIT@1 & HIT@10&HIT@1 & HIT@10& HIT@1 & HIT@10\\
\midrule
\multirow{3}{*}{\rotatebox[origin=c]{90}{\bf Sim}} &

NN EUC & 100& 100 &99.9&100& 66.5&73.7&99.9&100 \\
&NN COS&  100& 100 &99.8&100&67.3&74.6&99.8&100\\
&NN DOT& 0.73&2.06 &0.73&2.06&0.47&2.19&0.73&2.06\\

\midrule
\multirow{5}{*}{\rotatebox[origin=c]{90}{\bf Gradient}} &
IF & 0.01&0.34&0.01&0.35&0.04&0.25&0.01&0.35\\
&REP & 0.01&0.27&0.01&0.27&0.04&0.22&0.01&0.27\\
&RIF& 95.8 & 96.0&95.9&96.0&65.0&72.2&95.8&96.0\\
&GD&0.01&0.38&0.01&0.38&0.04&0.23&0.01&0.37\\
&GC&95.9&96.0&95.9&96.0&65.3&72.3&95.9&96.0\\
\bottomrule
\end{tabular}
\caption{Treating the training samples and their modifications as the target samples for attribution methods over the SST dataset.}
\label{tab:mod-SA}
\end{table*}
%%%%%%%%%%%%%%%%%%%
%%%%%%%%%%%%%%%
\begin{table*}
\centering
\begin{tabular}{lrrrrrrrrrr}
\toprule
&\multirow{2}{*}{\bf {Method}}&\multicolumn{2}{c}{\bf Train}&\multicolumn{2}{c}{\bf ADD}&\multicolumn{2}{c}{\bf Remove}&\multicolumn{2}{c}{\bf Replace}\\
\cmidrule(lr){3-4}
\cmidrule(lr){5-6}
\cmidrule(lr){7-8}
\cmidrule(lr){9-10}

& &HIT@1 & HIT@10& HIT@1 & HIT@10&HIT@1 & HIT@10& HIT@1 & HIT@10\\
\midrule
\multirow{3}{*}{\rotatebox[origin=c]{90}{\bf Sim}} &

NN EUC & 100& 100 &100& 100 &36.7&45.8 &100& 100\\
&NN COS&  100& 100 &100& 100 & 38.1&46.8&100& 100\\
&NN DOT&  1.30 & 6.44 &1.30 & 6.44 &3.49&10.7&1.30 & 6.44 \\

\midrule
\multirow{5}{*}{\rotatebox[origin=c]{90}{\bf Gradient}} &
IF &  0.0 & 0.01&0.0 & 0.01&0.02&0.10&  0.0 & 0.01\\
&REP & 0.0 & 0.01&0.0 & 0.01&0.01&0.09&0.0 & 0.01\\
&RIF& 92.5 & 92.5&92.5 & 92.5&32.6&41.2& 92.5 & 92.5&\\
&GD&0.0 & 0.01&0.0 & 0.01&0.10&0.50&0.0 & 0.01\\
&GC&92.5 & 92.5&92.5 & 92.5&32.8&41.2& 92.5 & 92.5&\\
\bottomrule
\end{tabular}
\caption{Treating the training samples and their modifications as the target samples for attribution methods over the MNLI dataset.}
\label{tab:mod-MNLI}
\end{table*}
%%%%%%%%%%%%%%%%%%%

\end{document}